\documentclass{article}
\usepackage{spconf,amsmath,epsfig}

\let\OLDthebibliography\thebibliography
\renewcommand\thebibliography[1]{
  \OLDthebibliography{#1}
  \setlength{\parskip}{0pt}
  \setlength{\itemsep}{0pt plus 0.3ex}
}

\begin{document}\sloppy

\def\x{{\mathbf x}}
\def\L{{\cal L}}
\title{FINED: FAST INFERENCE NETWORK FOR EDGE DETECTION}
\name{Jan Kristanto Wibisono and Hsueh-Ming Hang}
                   \address{Department of Electronics Engineering, National Chiao Tung University,Taiwan \\
                     jankristanto.03g@g2.nctu.edu.tw, hmhang@nctu.edu.tw}
\maketitle
\begin{abstract}
In this paper, we address the design of lightweight deep learning-based edge detection. The deep learning technology offers a significant improvement on the edge detection accuracy. However, typical neural network designs have very high model complexity, which prevents it from practical usage. In contrast, we propose a Fast Inference Network for Edge Detection (FINED), which is a lightweight neural net dedicated to edge detection. By carefully choosing proper components for edge detection purpose, we can achieve the state-of-the-art accuracy in edge detection while significantly reducing its complexity. Another key contribution in increasing the inferencing speed is introducing the training helper concept. The extra subnetworks (training helper) are employed in training but not used in inferencing. It can further reduce the model complexity and yet maintain the same level of accuracy. Our experiments show that our systems outperform all the current edge detectors at about the same model (parameter) size.
\end{abstract}
\section{Introduction}
\label{sec:Introduction}
Edge detection is an essential tool in image processing and computer vision. It has enormous applications in various computer vision tasks, such as scene understanding, object proposal, object detection, and image segmentation. In a typical image processing textbook, edges are lines separating two regions of significant intensity difference.

In the past a few years, numerous deep learning (DL) based edge detection methods have been proposed. And their accuracy is superior to that of the previous generations of edge detectors. However, the recent DL-based edge detectors are mostly based on the popular deep learning backbones with a huge number of parameters such as VGGNet~\cite{R1}. From the viewpoint of traditional edge detection methods (for example, gradient operators), these networks are overly complicated; they are slow in running time and have a massive number of parameters. Edge detection is a low-level image processing task and is often used as the base layer of high-level vision tasks. Thus, we believe edge detectors can be small and fast.

This paper is addressing the issue of model efficiency in DL-based edge detection. First, we propose a dedicated Edge Detection Network that has a small model and can provide good accuracy performance. Because of its low computational complexity in the inference phase, we call it, FINED, Fast Inference Network for Edge Detection. Second, we propose a specific additional structure in the training phase, called training helper. The training helper can train a high-performance backbone model used in the testing (inferencing) phase, but itself is eliminated in inference.
\section{Related Work}
\label{sec:Related Work}
Edge detection has a rich research history. We group the prior edge detectors into three categories. The first group is the traditional edge detection methods. Because edges appear at locations with sudden intensity changes, a popular way of finding the maximum changes is to find the maximum of the first derivative, or the second derivative is zero. The Sobel~\cite{R2} detector calculates the gradient magnitude using a set of 3x3 filter kernels. The Canny scheme~\cite{R3} starts with a smoothing process, an image is convolved with the Gaussian kernel. Then, it computes the gradient, and then applies the non-maximum suppression (NMS) algorithm to pinpoint edge locations. The main shortcoming of the traditional approach is that they only focus on the local intensity changes but fail to identify and ignore certain texture that are consider non-edges. 

The second group adopts the machine-learning techniques in constructing an edge detector. Researchers began to construct the texture descriptors, handcraft features, and sparse representation by combining the color, brightness, and other cues; examples are ~\cite{R4,R5,R6}. The learning-based edge detectors appear to overcome partially the texture problem that cannot be handled by the traditional approaches. Most published papers extract the hand-craft features and then use the classifiers trained with these features to identify edges. 

Along with the popularity of the convolutional neural network (CNN), the third group of edge detectors was proposed in the past a few years. Thanks to CNN’s ability to learn and produce robust features, it is also capable in detecting edges. The HED ~\cite{R7} is one of the first known DL-based edge detection, which uses VGGNet ~\cite{R1} together with a neural network classifier. HED fuses all the side-outputs from VGGNet to minimize a modified cross-entropy loss. Since then, there are many extensions based on HED and VGGNet, such as CED ~\cite{R8}, RCF~\cite{R9}, LPCB ~\cite{R10}, and BDCN ~\cite{R11}.  

Although CNN is very successful in many applications, it often has very high complexity. Hence, one recent trend is to design an efficient CNN structure with less computation and storage but can maintain the performance of a complex system. Inspired by the traditional edge detectors, Wibisono and Hang~\cite{R12} proposed a rather simple CNN system that can produce quite good results with a very small model. It is called TIN, Traditional Inspired Network ~\cite{R12}. In this paper, we like to further improve the accuracy of the proposed edge detector. Inspired by the concept in DSNet ~\cite{R13} that an additional sub-network may be used in the training process to produce a better-performance backbone network but this sub-network can be discarded in the inference phase to reduce the model size and to increase the speed. We call this sub-network, training helper. That is, we use multi-stage outputs in training and optimize them using the ground truth. The additional sub-networks help the entire system to converge faster in training, and they help to produce a strong backbone network to produce robust features for edge detection. During the inference phase, we remove the helpers, and the final output results are about the same as using the entire training system in inference.
\section{Fast Iinference Network}
\label{sec:Fast Iinference Network}
We propose a family of deep learning-based architecture for edge detection. We call it Fast Inference Network for Edge Detection, or FINED in brief. We adopt the dilated convolution ~\cite{R14} and Residual Relu ~\cite{R15} that can enlarge the receptive field and/or can enrich the features without adding many additional parameters. The Residual Pooling module is also included in our system, as hinted by RefineNet ~\cite{R15}. 
\subsection{Network Architecture}
TIN ~\cite{R12} has successfully demonstrated that a lightweight neural network can have a good performance on image edge detection. Furthermore, we extend the TIN architecture to strengthen its detection accuracy. TIN consists of Feature Extractor blocks together with Feature Enrichment and Summarizer operation. We modify the Feature Enrichment module, and combine it with the Residual Relu operator. We also modify the Summarizer using the Residual Pooling. Then, we match all the stage outputs to the ground-truth. After extensive study, we found that the multi-stage outputs can serve as a training helper that guides the training process to produce a higher performance system. In the inferencing phase, we remove the training helper to reduce the system complexity. 

\begin{figure}
\centering
\includegraphics[width=80mm]{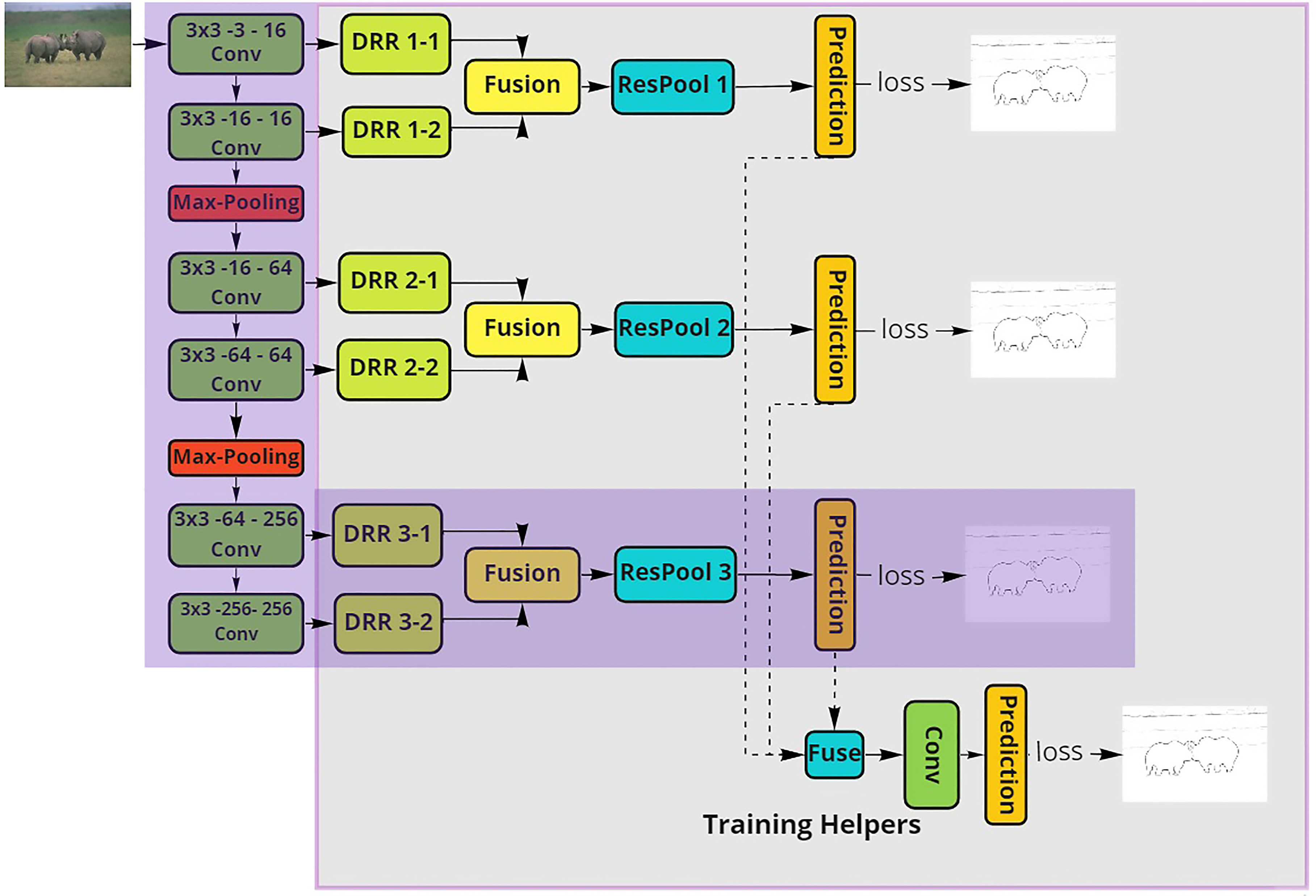} 
\caption{Overview of FINED3-Train Architecture}
\label{fig:F1}
\end{figure}

Figure~\ref{fig:F1} shows the overall proposed architecture for FINED3 (3-stage FINED system). The subnetworks with gray background are training helpers, which are active only during the training phase; in contrast, the subnetwork with purple background is active in both training and inference phases. We also propose FINED2 as a reduced version of FINED3. The difference is only the number of stages. FINED2 consists of only two stages.

The first column (3x3-3-16) in Figure~\ref{fig:F1} is the backbone network for feature extraction. It is made of mainly the 3x3 regular convolutional layer, and 3 input channels and 16 output channels. One stage is made up by two convolutional layers. To mimic the pyramid structure, we stack up two or three stages. The dimension (height x width) of output feature maps decreases as the stage number goes higher, and the feature channel number (number of filters) increases. The channel numbers are 16, 64, and 256, respectively, for stages 1, 2, and 3.

\begin{figure}
\centering
\includegraphics[width=80mm]{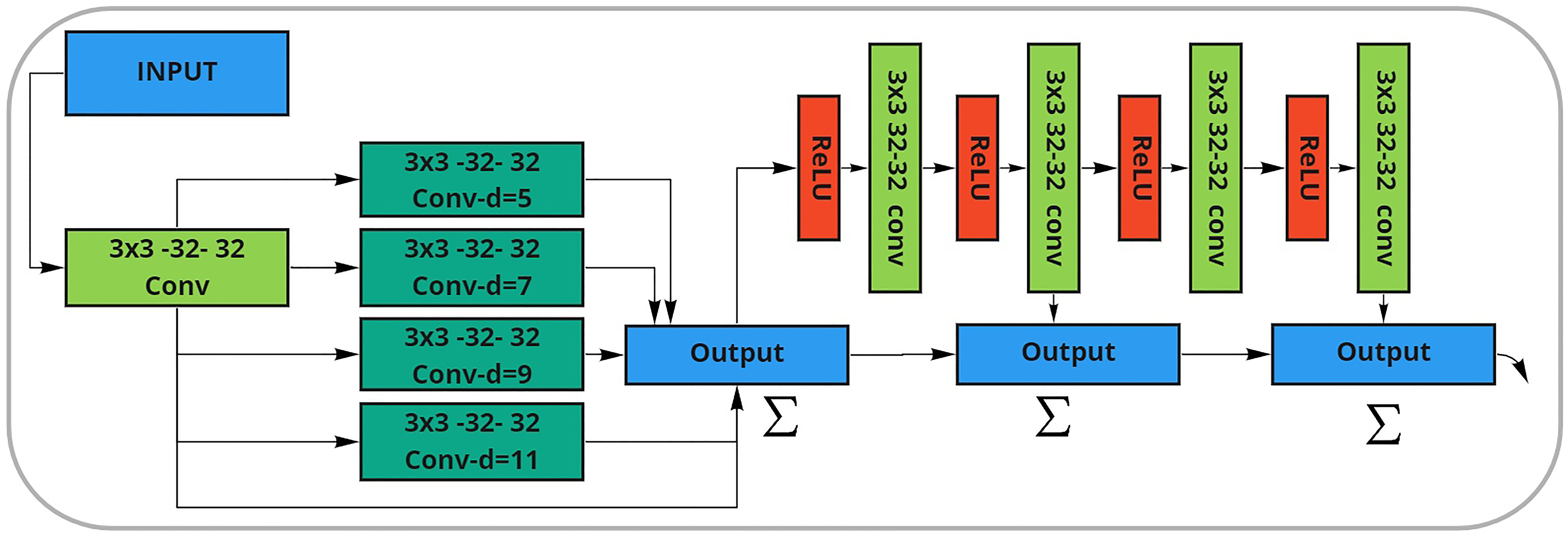} 
\caption{Dilated Residual Relu (DDR) Module}
\label{fig:F2}
\end{figure}

The Dilated Residual Relu (DRR) is added to enhance the Feature Enrichment module adopted in TIN. Figure~\ref{fig:F2} shows the design of our DRR module. Every feature extraction layer is connected to the Dilated Residual Relu (DRR) operator. Inside the DRR, we first use a 3x3 regular convolutional layer to accept the backbone net output features. The next layer is an aggregation of dilated convolutions. To capture different levels of receptive fields, we use different dilation sizes. The first dilation is 5, followed by 7, 9, and 11, and all the layers have 32 filters. At the end, we use pixel-wise aggregation to fuse DRR at each stage, but in order to reduce the feature channels from 32 to 8 before aggregation, we insert one regular convolutional layer.

The last module is Residual Pooling (ResPool). Inspired by RefineNet~\cite{R15} and Residual Block ~\cite{R16}, we include the average pooling and the regular convolutional layer and aggregate features of different scales to recognize objects of both small and large sizes. Because texture edges are often not counted as “edges”, we hope this structure can reduce the background texture edges and object texture edges and can capture only the boundaries of objects.

\begin{figure}
\centering
\includegraphics[width=80mm]{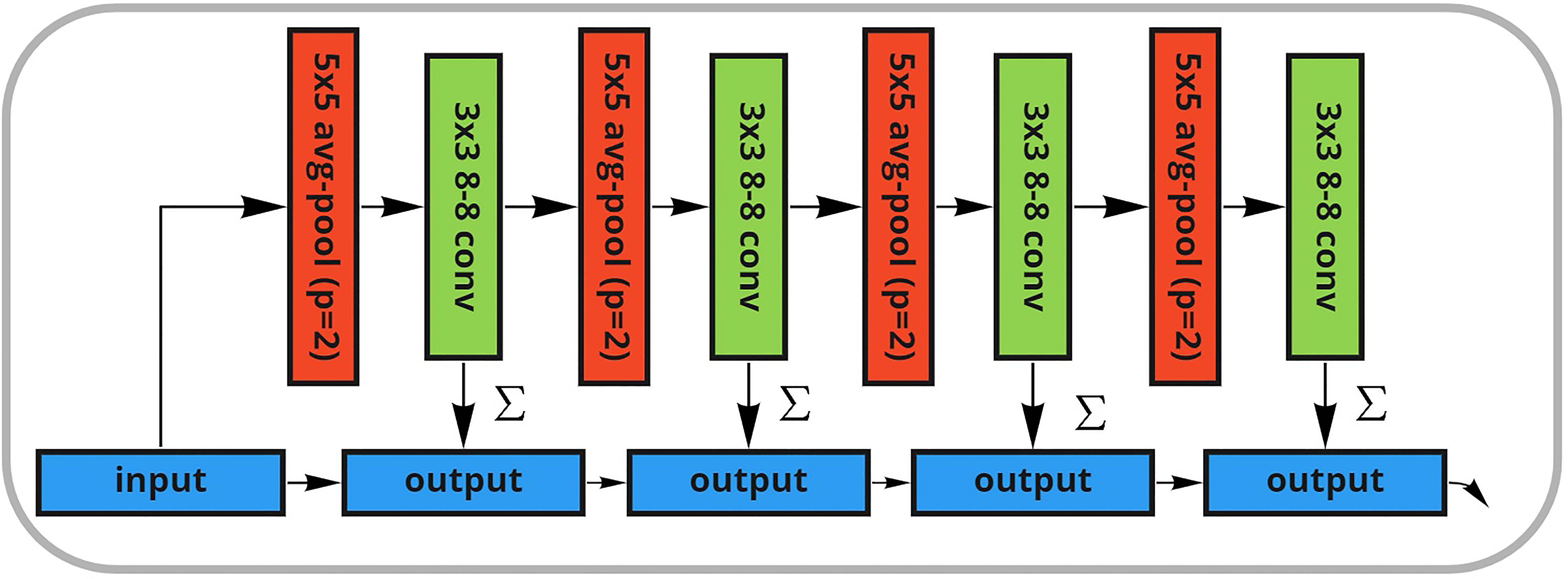} 
\caption{Residual Pooling Module}
\label{fig:F3}
\end{figure}
\subsection{Training Helpers}
In training, we have an output from each stage and have one additional fused output, which is a combination of all stage outputs. To create the fused output, we use the standard bilinear interpolation to upsample the low-resolution features. Then, we concatenate all the stage outputs together to form the fused output. In training, we minimize the aggregated cost function (against the ground-truth). On the other hand, only the last stage output is used in inferencing. The lower-level stage outputs (including the fused outputs) are used only in the training process and thus are called training helpers. They increase the training convergence speed and produce more robust features as shown in the ablation study (Section 4). In a way, the training helper plays the role of regulation to guide the backbone training.
\subsection{Loss Function}
The ground-truth data can sometimes be fuzzy. For example, in the BSDS500 dataset, several human annotators give several different annotations (labels) on edges. Although these annotations are not identical, they are often quite consistent. In the case of BSDS500, annotators give 0 to a non-edge pixel and 1 to an edge pixel. We take the average of all annotations to produce edge map ground-truth (probability values). Hence, our edge map ground-truth value ranges between 0 and 1. We ignore some weak edges because it may confuse the network. We remove weak edges by setting a threshold (confidence).

In training our architecture, the output of every stage and the fused output should match the ground truth. Hence, we define a loss function of each stage as follows.
\begin{equation}
  L(X;Y) = \sum_{i=1}^{I}  l(p_i;y_i)
  \label{equ:eq1}
\end{equation}
Where $X$ is the training image, $Y$ is the corresponding ground truth edge map, and $Y=\{y_i, i=1,…,I\}, y_i,:$ ground truth edge map pixel $i$, and $I:$ pixel number of $Y$. $P=\{p_i, i=1,…,I\}$ is the predicted edge map generated by the edge detector, $D(X;W)$, with weights $W$ for the input image $X$ is the input image.  In our system, the network in training, $D(X;W)$, can be different from the network in inferencing, $D_I(X;W)$. $l(p_i,y_i)$ denotes the chosen loss function for each stage output, which is specified by Eqn(2). In a typical image, about 10\% of pixels belonging to edges. To balance the edge and non-edge classes in loss function evaluation, inspired by ~\cite{R7,R9,R12}, we use $\alpha$ and $\beta$ as class-balancing weights in Eqn(2). In addition, we also use a threshold to ignore the weak edges. 
   \begin{equation}
    l(p_i;y_i)=
    \begin{cases}
      \alpha . log(1-\sigma(p_i)), & \text{if}\ y_i=0 \\
      0, & \text{if}\ 0 <  y_i < th \\
      \beta . log(\sigma(p_i)), & \text{otherwise}
    \end{cases}
    \label{equ:eq2}
  \end{equation}
Here, $\sigma()$ is the sigmoid function, and $th$ is a threshold of value 0.25 in our case. The values of $\alpha$ and $\beta$ are defined by Equation 3.
 \begin{equation}
	\alpha= \gamma . \frac{Y_+}{Y} \qquad \beta=\frac{Y_-}{Y}
	\label{equ:eq3}
 \end{equation}
 $Y_+$ and $Y_-$ are the numbers of pixels belonging to edge and non-edges, respectively. And, $Y=Y_+  + Y_-$. Parameter $\gamma$ is a hyper parameter, and we use 1.1 to emphasis more on the edges. 
Teh total loss function includes all the stage outputs (including the fused output). Thus, the total loss function is defined by Eqn 4, where $S$ is the number of stages. 
\begin{equation}
  LL(W;Y) = \sum_{s=1}^{S} \sum_{i=1}^{I} l(p_i^s;y_i) + \sum_{i=1}^{I} l(p_i^{fuse};y_i)
  \label{equ:eq4}
\end{equation}
\section{Experiment and Discussions}
\label{Experiment and Discussions}
\subsection{Implementation details}
We implement our scheme using Pytorch. We also adopt the TIN~\cite{R12} pretrained weights (model) to initialize the first two stages of our backbone networks. The other parameters are initialized with Gaussian distribution with zero-mean and standard deviation 0.01. The learning rate starts from 0.01, and then it is updated using a linear scaling factor, multiplying 0.1 for every ten epochs. And the batch-size is 3. The optimizer is stochastic gradient descent, and the training process terminates at 60 epochs. We conduct all the experiments on a single GPU, NVIDIA GeForce 2080Ti with 11G memory.
\subsection{Dataset}
In order to have a fair comparison with the other published works ~\cite{R6,R7,R9,R11,R12}, we evaluate our proposed network on the same Berkeley Segmentation Dataset (BSDS500) ~\cite{R4}. BSDS500 consists of 200 training, 100 validation and 200 test images. These 200 test images are our testing set. We combine the 200 training images and 100 validation images as the training set. We adopt the data augmentation technique same in RCF ~\cite{R9}. In addition, similar to RCF ~\cite{R9}, we also added PASCAL VOC ~\cite{R17} dataset and its flipped images into our training set.
\subsection{Evaluation}
Several previous edge detection publications already show that the multiscale testing procedure produce better performance ~\cite{R9,R10,R11}. We rescaled the inputs at three different scales [0.5, 1.0, 1.5] and fed them to the network. Then, we resize these outputs back to the original size and average these three outputs to get the final prediction. In addition, we use the Non-Maximum Suppression (NMS) as our post-processing for thinning the edge prediction maps. 

The detection accuracy is evaluated by the popular F-Measure with ODS and OIS. The ODS is Optimal Dataset Scale referring to the case that we set a threshold for all images in the testing set, whereas the IOS stands for Optimal Scale Image referring to the case we choose one threshold for each image. We set a tolerance parameter which controls the tolerance allowed to match the prediction and ground truth, which is 0.0075 for BSDS500. 
\subsection{Ablation study}
We train FINED2 and FINED3 on the BSDS500 training set together with PASCAL VOC, and evaluate various models on the BSDS500 test set. We study the impact of each component on the final result. We use ODS F-measure and the size of model parameters as our criteria to measure the system performance. A higher ODS indicates a better accuracy, and a smaller model size often leads to a lower storage and computational complexity.

Our first experiment is to examine the effectiveness of DRR and ResPool. DRR is the module right after the feature extraction backbone network at each stage. Table 1 shows that DRR improves the performance of both ODS and OIS compared to the original Feature Enrichment module ~\cite{R12} in TIN. With the additional ResPool, it can further enhance the performance. We conducted the experiments on both FINED2 and FINED3 architectures, and they show consistent results.

Second, we check the effectiveness of the training helper. We compare two architectures in inference. The entire FINED net is used in both training and inference is called FINED-Train. FINED-Inf architecture is the FINED-Train architecture without the training helper. For example, FINED3-Inf is the FINED3-Train architecture with inactive training helper; it is the purple portion of Figure~\ref{fig:F2}.
\begin{table}[t]
\begin{center}
\caption{Comparative studies on the FINED architectures. FINED2 refers to the FINED 2-stage training network. ‘w/-DRR’ refers to our proposed method with DRR active. ‘w/-RP’ refers to Residual Pooling active. ‘w/-FE’ refers to Feature Enrichment ~\cite{R12}.} \label{tab:tabl1}
\begin{tabular}{|c|c|c|c|}
  \hline
  Method & ODS & OIS & \#P (million)\\
\hline
FINED2-w/-DRR-w/-RP  & 0.779 & 0.800 & 0.396\\
FINED2-w/-DRR-w/o-RP  & 0.775 & 0.796 & 0.391\\
FINED2-w/-FE-w/-RP	& 0.774 &	0.794 & 0.24\\
FINED2-w/-FE-w/o-RP  & 0.772 & 0.795 & 0.24\\
FINED3-w/-DRR-w/-RP  & 0.790 & 0.808 & 1.43\\
FINED3-w/-DRR-w/o-RP  & 0.784 & 0.802 & 1.42\\
FINED3-w/-FE-w/-RP &0.786 &0.805 &1.21\\
FINED3-w/-FE-w/o-RP  & 0.780 & 0.799 & 1.20\\
\hline
\end{tabular}
\end{center}
\end{table}
Table 2 shows the impact of training helpers (TH) in our architecture. We compare FINED3-Train and FINED3-Inf. We also tried that if the FINED3-Inf is trained without the pretrained weights obtained from the FINED3-Train model. In Table 2, FINED3-Inf-w/-TH refers to the FINED3-Inf architecture, but we use the model (weights) obtained from FINED3-Train. And the FINED3-Inf-w/o-TH refers to the case that we train the FINED3-Inf network without using the model weights derived from FINED3-Train. It is shown in Table 2 that FINED3-Inf-w/o-TH is much worse than FINED3-Inf-w-TH. We did this set of experiments for FINED3 and FINED2 architectures, and both show consistent results. The training helpers indeed produce an inference model that achieves much better results. At the same time, it reduces significantly (30\%) the model size in inferencing.
\begin{table}[t]
\begin{center}
\caption{Ablation studies of FINED-Inf using training helper (TH) and without training helper.} \label{tab:tabl2}
\begin{tabular}{|c|c|c|c|}
  \hline
  Method & ODS & OIS & \#P (million)\\
\hline
FINED2-Inf-w/-TH  & 0.779 	&0.799&	0.23\\
FINED2-Inf-w/o-TH  & 0.756&	0.778&	0.23\\
FINED2-Train	& 0.779&	0.800	&0.39\\
FINED3-Inf-w/-TH  & 0.788&	0.804&	1.08\\
FINED3-Inf-w/o-TH  & 0.773&	0.791&	1.08\\
FINED3-Train  & 0.790&	0.808&	1.43\\
\hline
\end{tabular}
\end{center}
\end{table}
In our experiments, we observed the following points. The evaluation metric shows that FINED-Train and FINED-Inf produce similar results. Second, if we continue to retrain the inference architecture (FINED-Inf), it does not improve the results. 
\begin{table}[t]
\begin{center}
\caption{Ablation studies of FINED3 using VGGNet as backbone.} \label{tab:tabl3}
\begin{tabular}{|c|c|c|c|}
  \hline
  Method & ODS & OIS & \#P (million)\\
\hline
FINED3-Train-VGG &	0.796 &	0.817&	1.86\\
FINED3-Inf-VGG&	0.795&	0.811&	1.44\\
\hline
\end{tabular}
\end{center}
\end{table}
We further extend our architecture by enlarging the backbone filter numbers inspired by VGGNet. It simply changes the number of filters from 16 to 64 and 64 to 128 for stage 1 and stage 2, respectively. During training, we initialize the backbone using the VGGNet (stage 1-3) pretrained weights. 

\begin{figure}
\centering
\includegraphics[width=80mm]{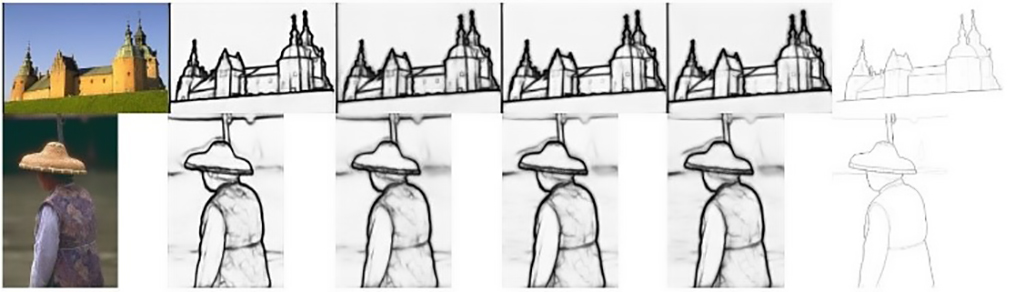} 
\caption{Comparison among the proposed methods. From left to right: input, FINED3-Train, FINED3-Inf, FINED3-Train-VGG, FINED3-Inf-VGG, ground-truth}
\label{fig:F4}
\end{figure}

We evaluate the effects of this simple modification and summarize experimental results in Table 3. it can be observed that, by enlarging the backbone, it boosts the baseline performance, from 0.79 to 0.796 in the ODS F-measure. It not only improves the training architecture but also improves the inference architecture from 0.788 to 0.795 in the ODS F-measure. The 0.7\% improved performance is at the cost of an increase of about 0.3M~0.4M parameters. Figure~\ref{fig:F4} shows the visual comparison between FINED3 and FINED3-VGG on both inference network and training network. As we can see that FINED3-VGG produces less visual noise and can reduce some texture edges compared to the original FINED3. The FINED3-Train-VGG and FINED3-Inf-VGG networks have similar performance on both edge maps and the evaluation metrics. This again proves the concept of training helper, which is reliable and works for a variety of architectures. 
\section{Experimental Results}
\begin{figure}
\centering
\includegraphics[width=80mm]{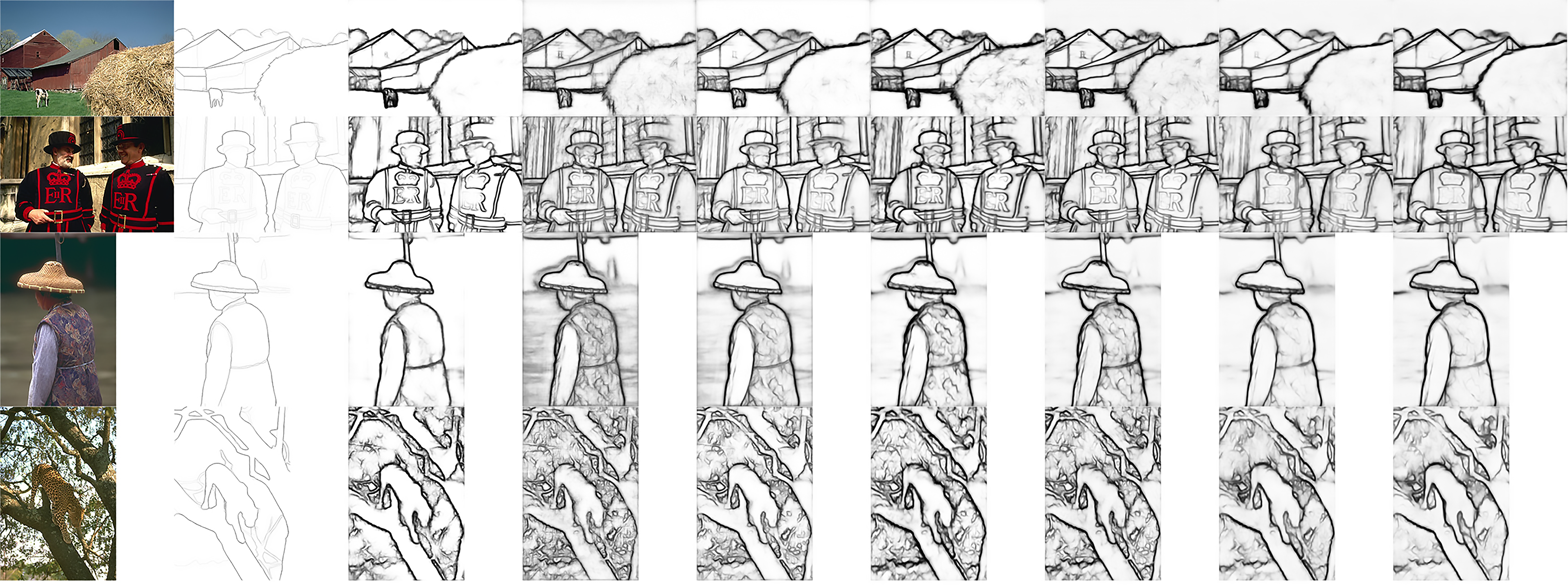} 
\caption{Visual comparison among edge detectors on the BSDS500 test set. From left to right: input, ground-truth, HED, BDCN2, BDCN3, TIN2, FINED2-Inf, FINED3-Inf, FINED3-Inf-VGG}
\label{fig:F5}
\end{figure}

\begin{figure}
\centering
\includegraphics[width=80mm]{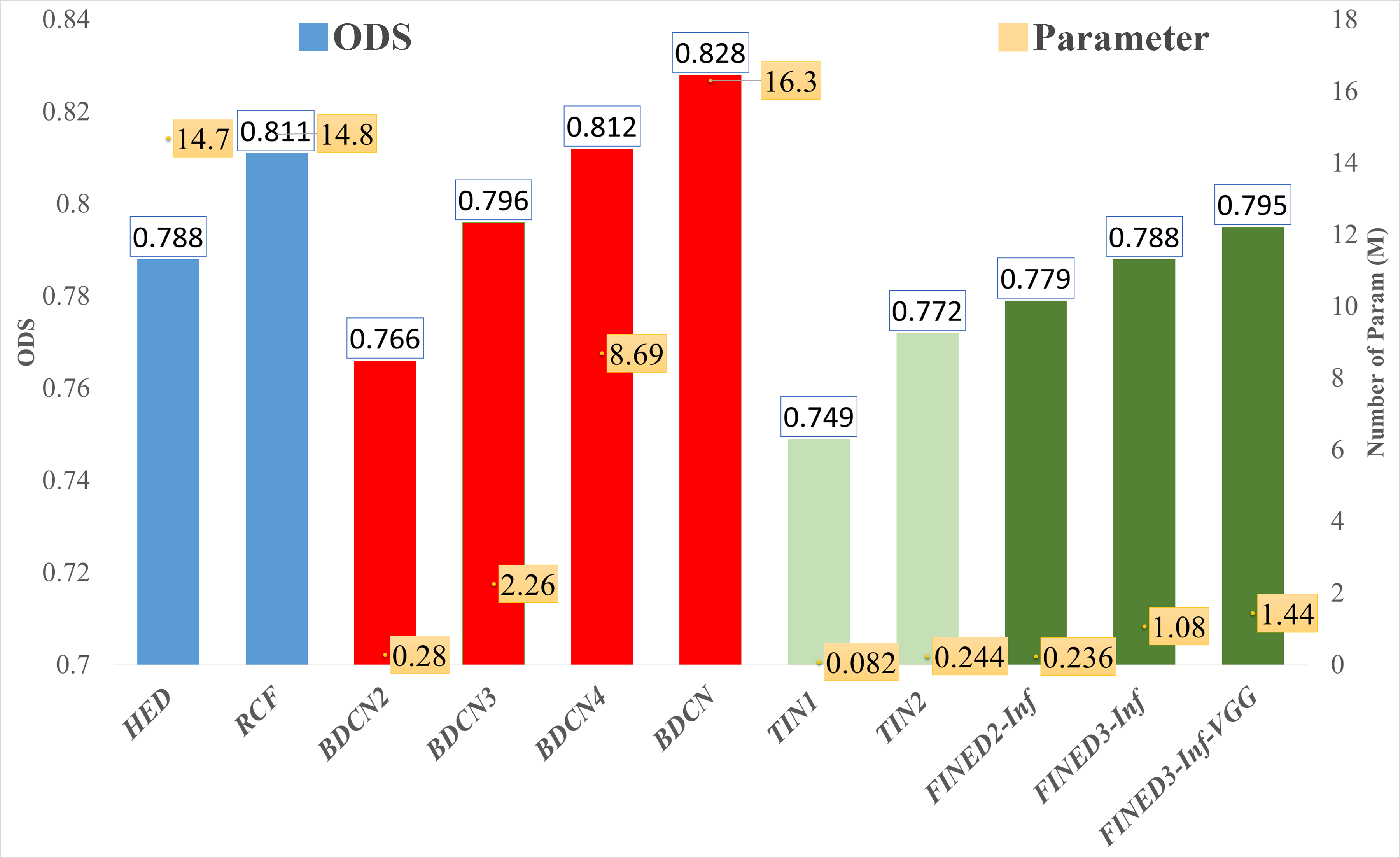} 
\caption{Comparison of complexity and performance with the other approaches }
\label{fig:F6}
\end{figure}

We compare our proposed systems with the other learning-based edge detectors. The classical approach, Canny~\cite{R3}, and two machine learning approaches, gPb-UCM~\cite{R4} and SF~\cite{R6} are included as reference points. The popular deep learning (DL) based edge detectors in Table 4 are HED ~\cite{R7}, CED~\cite{R8},  RCF~\cite{R9}, LPCB~\cite{R10}, BDCN~\cite{R11}, and TIN~\cite{R12}. The visual results are shown in Figure~\ref{fig:F5}. And Figure~\ref{fig:F6} summarizes their accuracy metric and model size. 

So far, BDCN is the leading edge detector in accuracy. But, our goal is to design lightweight edge detectors. As we can see in Table 4, our methods, both FINED2 and FINED3, outperform the classical and machine learning edge detectors. At comparable or smaller model size, FINED2 and FINED3 have the best accuracy among the DL-based edge detectors. FINED2-Inf achieves 0.779 ODS; it outperforms BDCN2 and TIN2 with a smaller number of parameters. Our FINED3-Train-VGG and FINED3-Inf-VGG have about the same accuracy with BDCN3, but with a smaller model. In addition, we also compare the inference speed. FINED2-train runs at about 160fps and FINED2-Inf at about 203fps. This running speed comparison shows that the training helpers work well on decreasing the complexity and boosting the inference accuracy and speed.
\begin{table}[t]
\begin{center}
\caption{Comparison of various edge detectors tested on the BSDS500 dataset.} \label{tab:tabl4}
\begin{tabular}{|c|c|c|c|c|}
  \hline
  Method & ODS & OIS & \#P & FPS\\
\hline
Canny~\cite{R3}	&0.611&	0.676& & \\
gPb-UCM~\cite{R4} &	0.729&	0.755& & \\
SF~\cite{R6}	&0.743&	0.763& & \\		
HED~\cite{R7}&	0.788&	0.808&	14.7&	\\
CED~\cite{R8}&	0.815&	0.834&	21.4&	10\\
RCF~\cite{R9}&	0.811&	0.83&	14.8&	10\\
LPCB ~\cite{R10}&	0.815&	0.834&		&\\
BDCN2~\cite{R11}&	0.766&	0.787& 	0.48&	37\\
BDCN3~\cite{R11}&	0.796&	0.817& 	2.26&	33\\
BDCN4~\cite{R11}&	0.812&	0.830& 	8.69&	29\\
BDCN5~\cite{R11}&	0.828&	0.844&	16.3&	22\\
TIN1~\cite{R12}&	0.749&	0.772&	0.08&	\\
TIN2~\cite{R12}&	0.772&	0.792&	0.24&\\
FINED2-Train (ours)&	0.779&	0.8&	0.39&	160\\
FINED2-Inf (ours)	&0.779&	0.799&	0.23&	203\\
FINED3-Train (ours)&	0.790&	0.808&	1.43&	99\\
FINED3-Inf (ours)	&0.788&	0.804&	1.08&	124\\
FINED3-Train-VGG (ours)&	0.796&	0.817&	1.86&	\\
FINED3-Inf-VGG (ours)&	0.795&	0.811&	1.44&	\\
\hline
\end{tabular}
\end{center}
\end{table}
Figure~\ref{fig:F5} shows the visual comparison among several recent edge detection systems and FINED on the BSDS500 test set. Compared to BDCN2 and BDCN3, although our method has similar ODS, our edge maps seem to be visually slightly better. Figure~\ref{fig:F6} is a summary of the complexity and the accuracy of several deep learning-based edge detectors. Our FINED2-Inf, with only 0.236M parameters, has a higher ODS F-measure than BDCN2 and TIN2, which have more parameters. Our FINED3-Inf with only 1.08M parameters can archive the ODS F-measure at 0.778, which is close to HED with 14M more parameters. FINED3-Inf-VGG has 1.44M parameters but it achieves an ODS of 0.795, similar to that of BDCN3 ~\cite{R11}, which has 2.26M parameters.
\section{Conclusion}
\label{Conclusion}
We propose a neural network architecture (FINED) dedicated to edge detection. The FINED family focuses on lightweight and fast architecture with good edge detection accuracy. FINED adopts dilated convolution and Residual ReLU to enrich the features with minimal additional model weights. In addition to that, FINED employs the Residual Pooling to remove the background context. The concept of training helper is proposed, and it shows good performance. The extra subnetworks (training helper) are employed in training to produce a good backbone system that can produce nearly the same accurate results in inferencing (without training helper). We thus can reduce the model complexity and maintain the accuracy performance at the same time. Our proposed schemes can produce good edges in a very efficient manner. FINED2-Inf outperforms BDCN2 with a much smaller model (parameter size). Our FINED3-Inf-VGG has the same accuracy as BDCN3 and able to reduce 36\% of its parameters.
\bibliographystyle{IEEEbib}
\bibliography{icme2021template}

\newpage

\begin{appendix}\sloppy
    
   \begin{center}
    \textbf{APPENDIX}
   \end{center}

\pagestyle{empty}
\def\x{{\mathbf x}}
\def\L{{\cal L}}
In this document, we provide additional material related to our paper, which includes multiscale testing, filters and feature map visualization. We also present additional qualitative results and analysis.
\section{MULTISCALE TESTING}
\label{sec:multiscale}
In this section, we discuss the effectiveness of the multiscale testing and its impact on the final results. In the ordinary inference, we only use one scale of the input image. In the one scale case, we feed the original input into the model. Then, the model produces the edge probability prediction. In the multiscale testing strategy, the inputs fed into the network have three different scales [0.5, 1, 1.5] of the same image. Then, we take the average of the outputs as the final prediction. Other than these three scales, we also try different scales, but we found that the above three scales are the optimum cases.

We evaluate the effect of multiscale testing on both FINED2 and FINED3. Although this multiscale testing is a simple process, it is able to improve ODS and OIS. Table ~\ref{tab:S1} shows that the multiscale testing improves the accuracy of all FINED models. This multiscale testing increases ODS from 0.784 to 0.788 for FINED3 and from 0.771 to 0.779 for FINED2. But this multiscale testing reduces the inference speed for an individual test. 

\section{FILTERS AND FEATURE MAPS}
\label{sec:filterandfeaturemap}
\begin{figure}
\centering
\includegraphics[width=80mm]{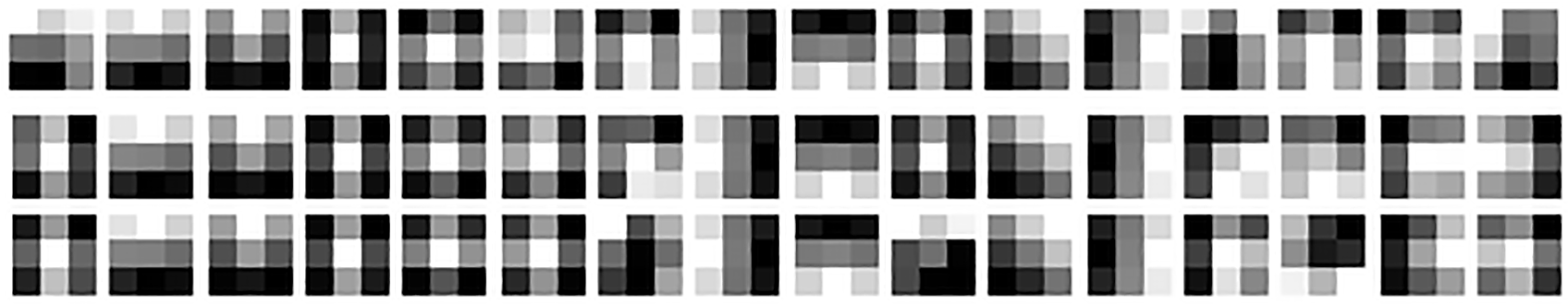} 
\caption{First convolutional layer filter of the FINED3-Train (FINED3-Inf) model}
\label{fig:S1}
\end{figure}

\begin{figure}
\centering
\includegraphics[width=80mm]{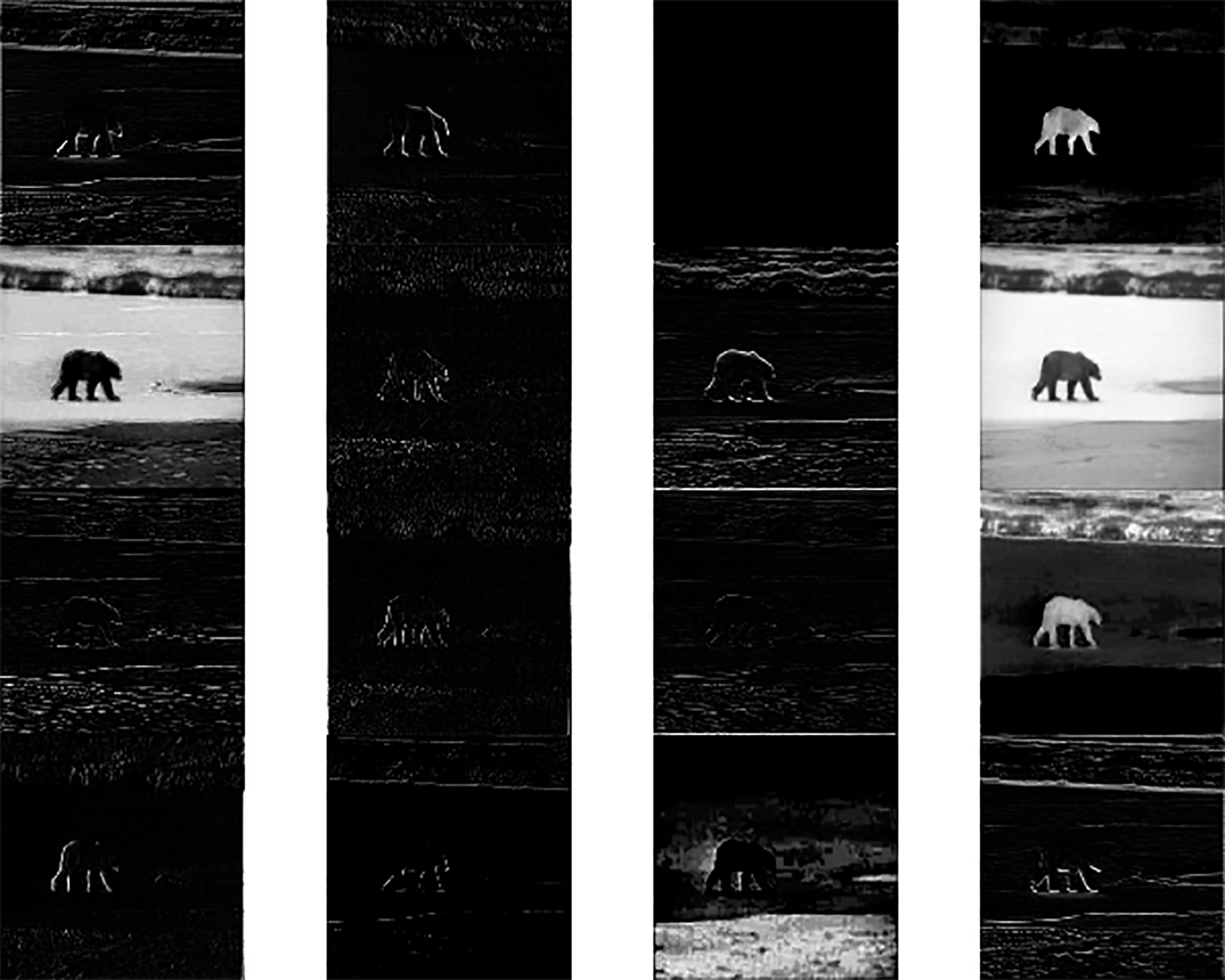} 
\caption{Feature maps at the output of the first convolutional layer (conv1\_1) of the FINED3-Train (FINED3-Inf) model}
\label{fig:S2}
\end{figure}

Neural networks have delivered several state-of-the-art solutions to the edge detection problem. Here, we like to investigate the process how a neural network decides a particular pixel to be an edge or non-edge. Due to the high nonlinearity nature of neural networks, the explainable theoretical modeling of a deep neural network is a topic still under intensive study. Hence, we try to answer this question by looking at some filters and feature maps. Figure ~\ref{fig:S1} shows the 3x3 filter weights (coefficients) at the first convolutional layer (conv1\_1 in the paper Fig.1) in the FINED3-Train (also, FINED3-Inf) architecture. The input is a three-channel image (RGB), and the convolution layer is 3x3 (filter size) with three input channels and 16 output channels. So, there are 3x16=48 filters, as shown in Figure ~\ref{fig:S1}. Observing the weight patterns, a large percentage of them are line and ramp detectors of various directions. This matches the early vision part of human visual system.
\begin{table}[t]
\begin{center}
\caption{Comparative studies of FINED-Inf using multiscale testing (MS) or not.} \label{tab:S1}
\begin{tabular}{|c|c|c|c|}
  \hline
  Method  & ODS & OIS & \#P  \\
  \hline
  FINED2-w/-MS&	0.779 & 	0.799	&0.23 \\
FINED2-w/o-MS&	0.771&	0.792&	0.23 \\
FINED3-w/-MS&	0.788&	0.804&	1.08 \\
FINED3-w/o-MS&	0.784&	0.800&	1.08 \\
  \hline
\end{tabular}
\end{center}
\end{table}
Figure ~\ref{fig:S2} shows the feature maps at the output of the first convolutional layer (conv1\_1). It contains 16 feature maps. In Figure ~\ref{fig:S2}, some maps show edge maps and some show object segmentation maps. It is interesting to see that not all feature maps are edges. An interpretation to this phenomenon is that the ground truth edges in the given dataset often focus more on the object boundaries. The background texture or even the texture on the foreground objects are often ignored in the ground truth maps. Hence, the network tries to detect “objects” to reduce the edges associated with the fine textures. The feature maps from stage 1 will be passed to stage 2 and stage 3.

\begin{figure}
\centering
\includegraphics[width=80mm]{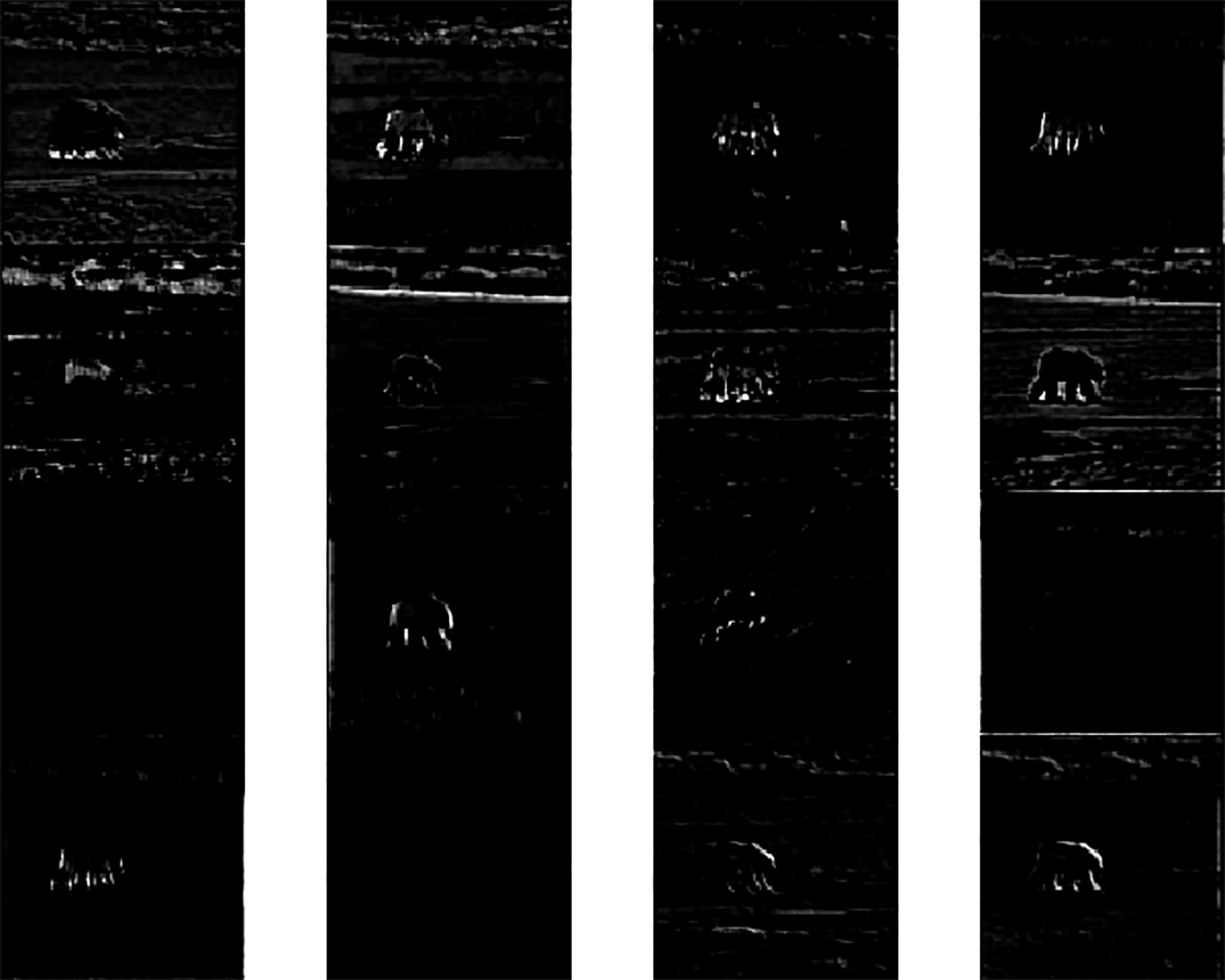} 
\caption{16 Feature Maps from Conv3 1 of the FINED3-Train (FINED3-Inf}
\label{fig:S3}
\end{figure}

\begin{figure}
\centering
\includegraphics[width=80mm]{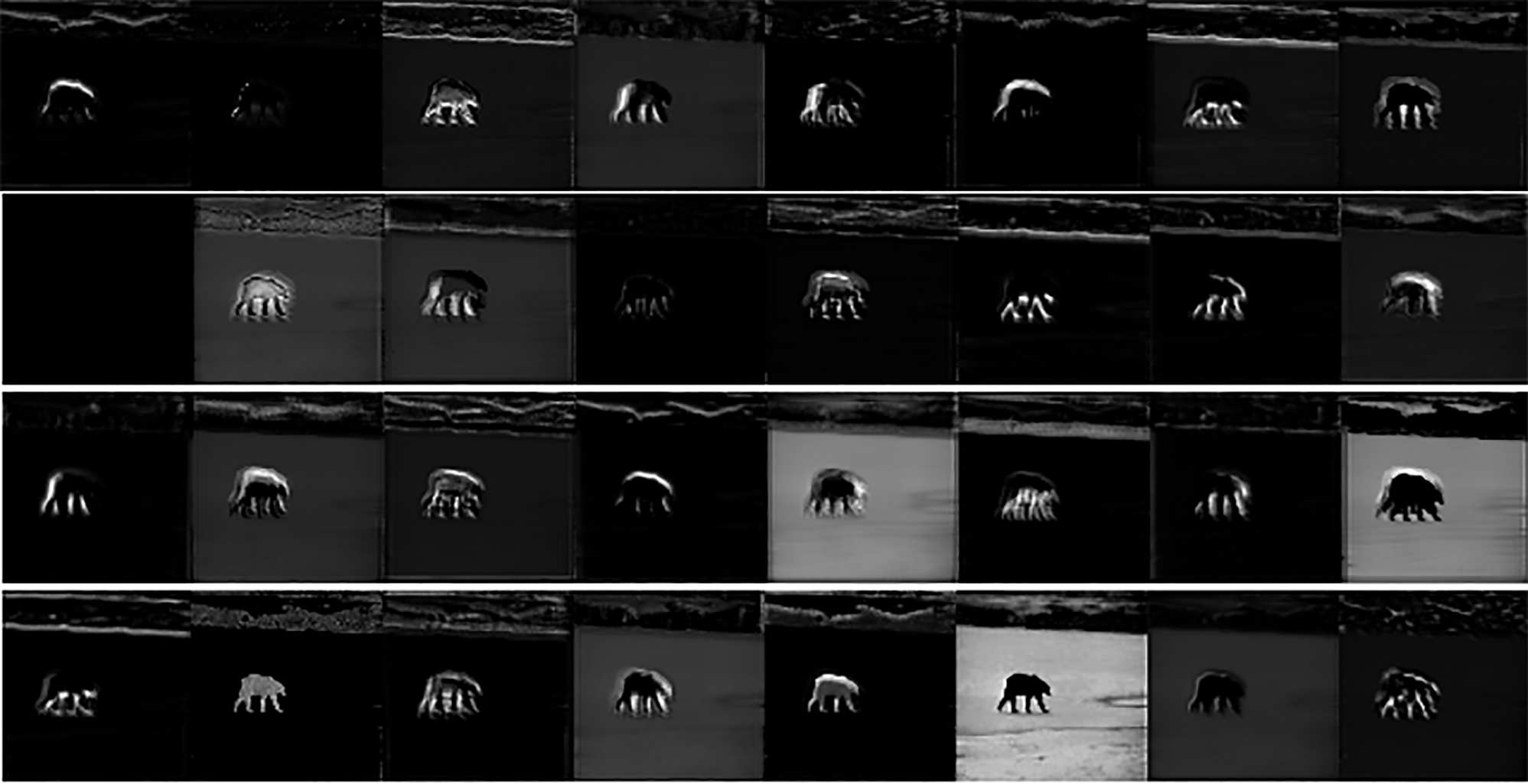} 
\caption{Feature maps at the output of DRR1 1 of FINED3-Train}
\label{fig:S4}
\end{figure}

\begin{figure}
\centering
\includegraphics[width=80mm]{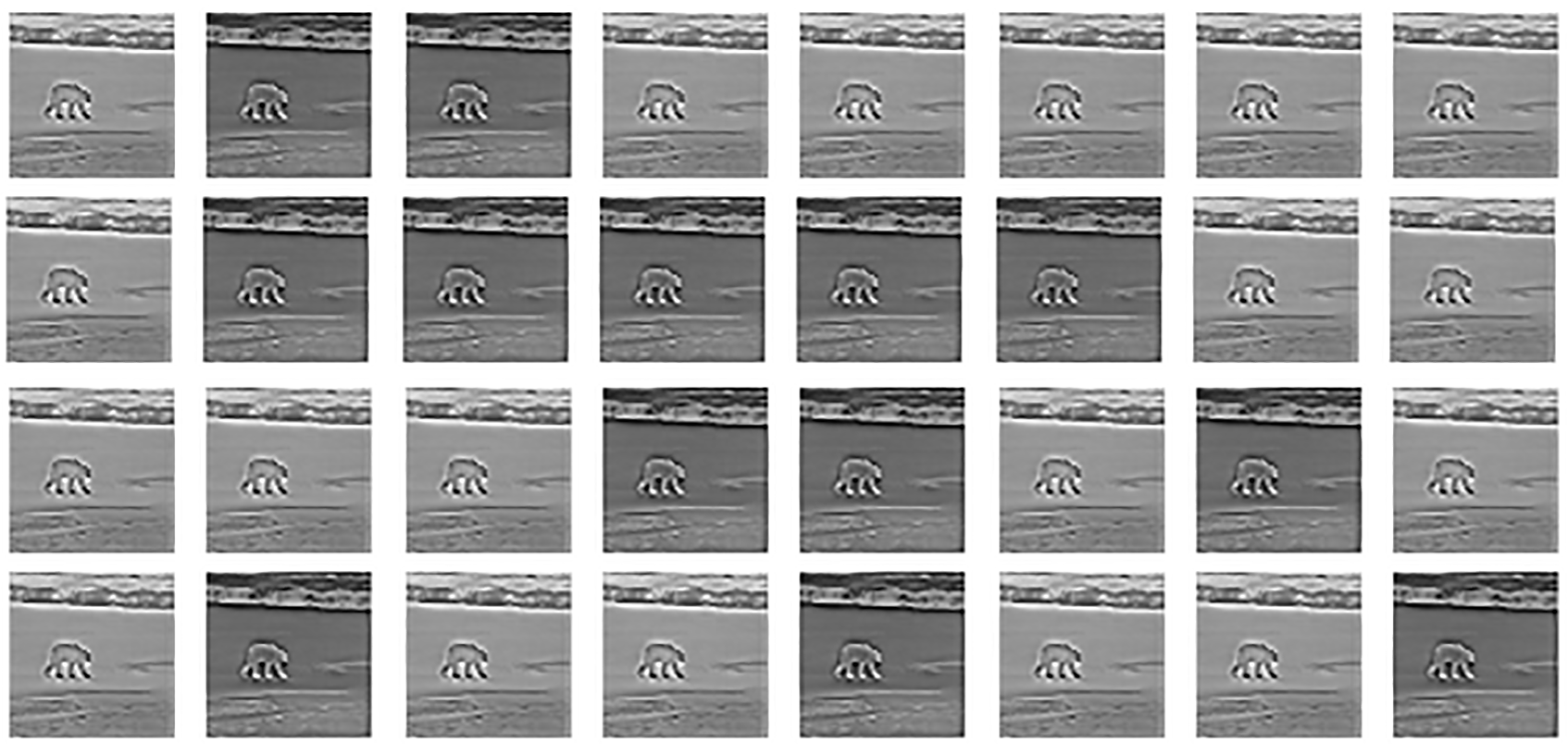} 
\caption{Feature maps from DRR3\_1of FINED3-Train (FINED3-Inf)}
\label{fig:S5}
\end{figure}

\begin{figure}
\centering
\includegraphics[width=80mm]{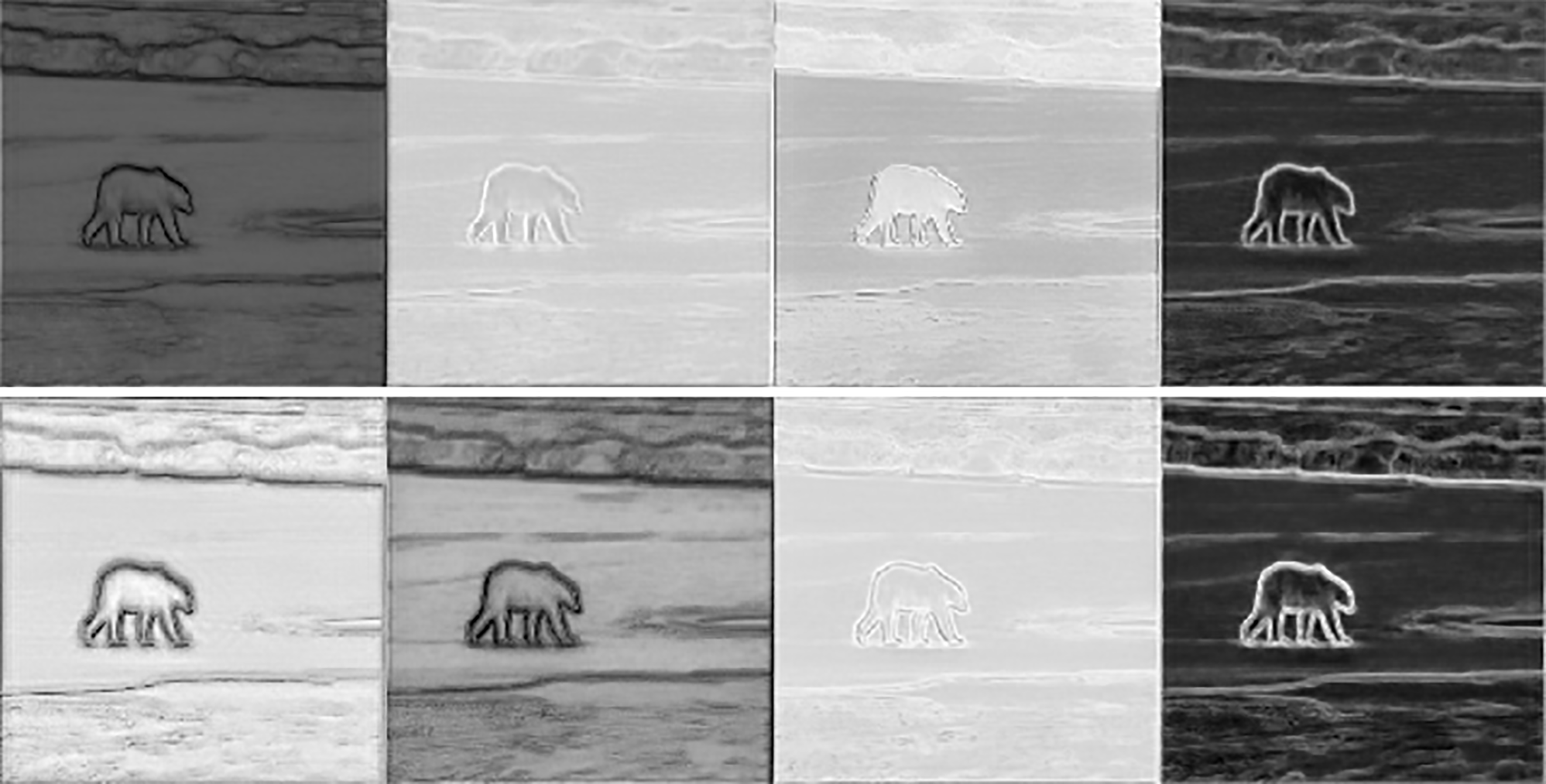} 
\caption{Feature maps at the output of Residual Pool 1 of FINED3-Train}
\label{fig:S6}
\end{figure}

\begin{figure}
\centering
\includegraphics[width=80mm]{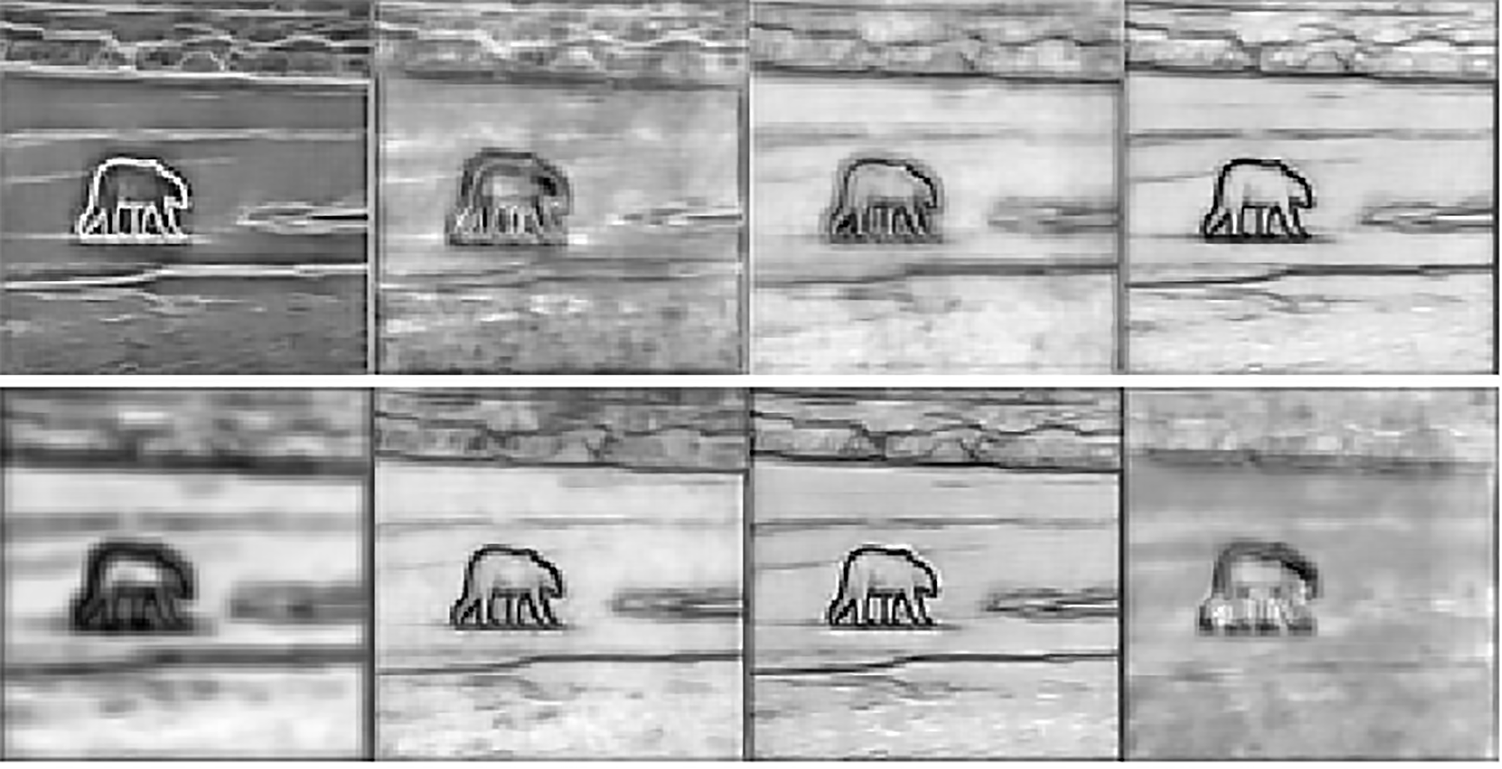} 
\caption{Feature maps from Residual Pool 3 of FINED3-Train (FINED3-Inf)}
\label{fig:S7}
\end{figure}

Conv3\_1 is one of backbone CNN layers for stage 3. There are 256 feature maps in the conv3\_1 outputs. Limited by space, we only display 16 randomly chosen features in Figure ~\ref{fig:S3}. Also note that the resolutions (dimensions) of these feature maps are 1/4 (in each dimension) of those in conv1\_1. Hence, the edges shown in these feature maps are actually narrower. As we can see, most of them are object edges while the texture edges are suppressed. Some feature maps are very dark but they still contain object/background information and that information can be extracted by the following CNN layers. The feature maps from the backbone network (conv1\_1, conv1\_2, . . . , conv3\_2) can be thought as the raw features. They are further processed and refined by the DRR modules. Figure ~\ref{fig:S4} shows the feature maps from the DRR1\_1 module. (DDR1\_1 exists in FINED3-Train not in FINED3-Inf.) These feature maps are obtained by convolving the feature maps in Figure ~\ref{fig:S1} and the DRR1\_1 filter weights.
As we can see in Figure ~\ref{fig:S4}, the network is able to extract various edge/object patterns from the inputs. There are several different types of patterns: some features highlight
the background, some highlight the objects, and one feature map often shows only a portion of the object edges, for example, upper edge of object, right-side edges of object, etc. For comparison, Figure ~\ref{fig:S5} shows the feature maps from the output of DRR3\_1. The feature map resolution is 1/4 of the counterpart at DRR1\_1. The edge patterns are starting to develop, since it is close to the end of the system. One layer closer to the final prediction layer, Figure ~\ref{fig:S6} shows the feature maps at the output of ResPool\_1 (FINED3-Train). Our system tries to reduce the difference between the neural network final outputs and the edge ground-truth. At the end, the network generates edges of the input image. This is particularly true for ResPool 3 (Figure ~\ref{fig:S7}), which is next to the final output. It is interesting to see that some feature maps of ResPool\_3 seem to blur the textures.

\begin{figure}
\centering
\includegraphics[width=80mm]{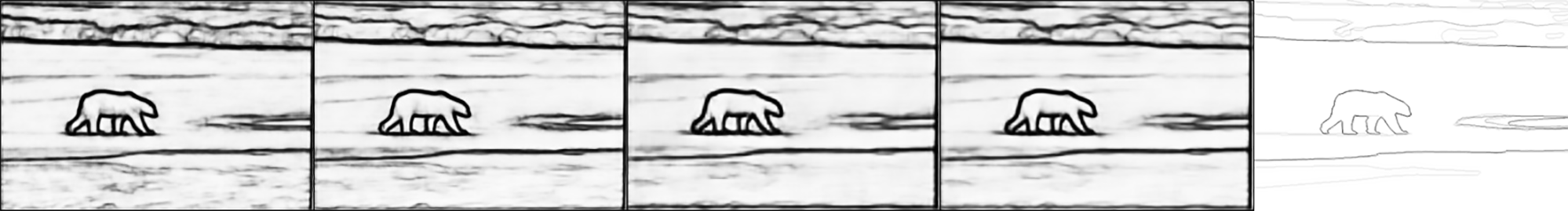} 
\caption{All output prediction maps of FINED3-Train model. Left to right: stage 1, stage 2, stage 3, and the fused output. The rightmost one is the groundtruth}
\label{fig:S8}
\end{figure}

The last layer in the system is a 3x3 convolutional layer. The role of this layer is to force the neural network to produce the probability map of the same size as the input; and the training target is to reduce the weighted cross-entropy. Figure ~\ref{fig:S8} shows all the outputs from the FINED3-Train architecture. As described in our main paper, the FINED3-Train has four outputs: 3 stage outputs, and one fused output. All of them are trained to match the ground-truth (GT) edges. And indeed, they are all very similar. But clearly, the stage 3 output and the fused output give the best result, and on the average, these two provide about the same accuracy. Therefore, we can simply use the stage 3 output as the final result in the inference (testing) phase. Taking this strategy can significantly reduce the model size.

\section{QUALITATIVE RESULTS}
\label{sec:quantitativeresults}
Figure ~\ref{fig:S9} shows 10 quantitative results (Precision-Recall curves) including HED ~\cite{R7}, BDCN2-MS ~\cite{R11}, BDCN3-MS ~\cite{R11}, TIN1 ~\cite{R12}, TIN2 ~\cite{R12}, and also our proposed architectures, FINED2-Train, FINED2-Inf, FINED3-Train, and FINED3-Inf on the BSDS500 dataset (model size in million, in parentheses). FINED2-Inf is our most successful architecture. So far, there is no architecture with the same or smaller complexity and have comparable performance, including the stage-of-the-art, BDCN2-MS. Figure ~\ref{fig:S9} shows that FINED3-Train outperforms the HED with a much lower (1.43M) complexity. The FINED3-Train-VGG achieves comparable performance (ODS=0.796) to the state-of-the-art method (BDCN3-MS) but using a smaller number of parameters (1.86M). To reduce the complexity while maintaining the accuracy, we introduce the training helper as described in the main paper. FINED3-Inf-VGG shows the robustness of the training helper. It achieves ODS=0.795, but successfully reduces 25\% of the model size. 

\begin{figure*}
\centering
\includegraphics[width=120mm,height=12cm]{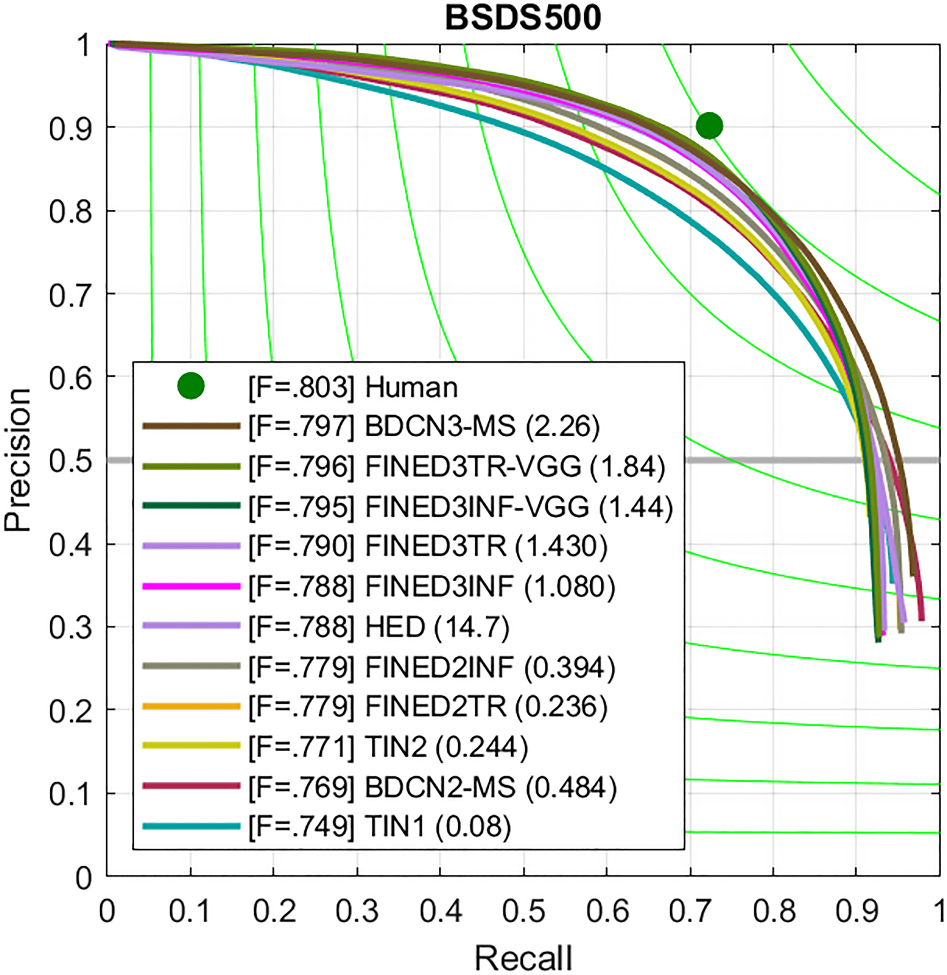} 
\caption{The PR curves of various systems on BSDS500}
\label{fig:S10}
\end{figure*}

\begin{figure*}
\centering
\includegraphics[width=140mm,height=20cm]{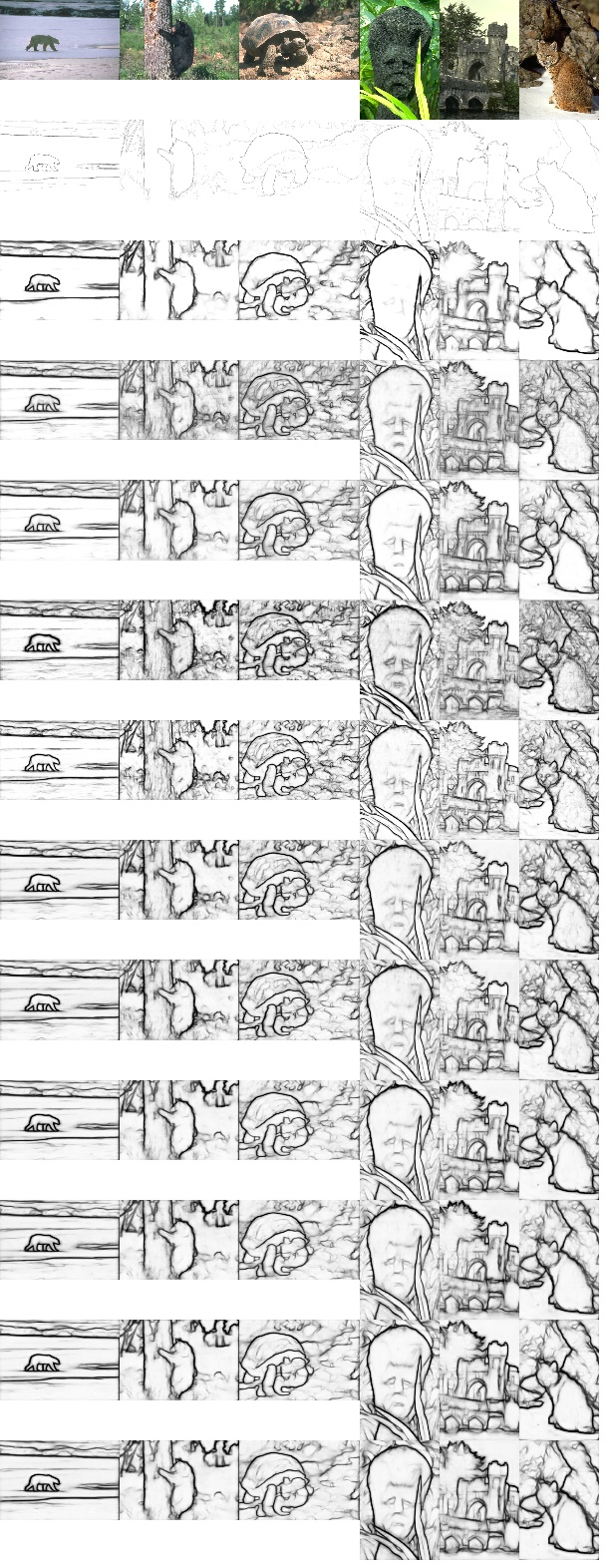} 
\caption{Subjective comparison on the BSDS500 dataset. Top to bottom: input, ground-truth, HED~\cite{R7}, BDCN2-MS~\cite{R11}, BDCN3-MS~\cite{R11}, TIN1~\cite{R12}, TIN2~\cite{R12}, FINED2-Train (ours), FINED2-Inf (ours), FINED3-Train (ours), FINED3-Inf (ours), FINED3-Train-VGG(ours), FINED3-Inf-VGG(ours).}
\label{fig:S9}
\end{figure*}

\end{appendix}

\end{document}